\documentclass[letterpaper, 10 pt, conference]{ieeeconf}  

\IEEEoverridecommandlockouts                              
                                                          
\usepackage{amsmath}
\usepackage{algorithmic}
\usepackage[caption=false,font=footnotesize]{subfig}
\usepackage{url}
\usepackage{hyperref}

\usepackage[noadjust]{cite} 
\usepackage{array}
\usepackage{booktabs}
\usepackage{xcolor}
\usepackage{amssymb}  
\usepackage{bm}  
\usepackage[export]{adjustbox} 
\usepackage[ruled,norelsize]{algorithm2e} 
\usepackage{pifont} 
\usepackage{multirow}

\def\ie{\emph{i.e.}} 
\def\eg{\emph{e.g.}} 

\DeclareMathOperator{\E}{\mathbb{E}}
\DeclareMathOperator*{\argmax}{arg\,max}

\definecolor{green(ryb)}{rgb}{0.4, 0.69, 0.2}
\definecolor{orange(ryb)}{rgb}{0.98, 0.6, 0.01}

\newcommand{\discvae}{DiSCVAE}

\title{Disentangled Sequence Clustering for \\ Human Intention Inference}

\author{Mark Zolotas and Yiannis Demiris
\thanks{M. Zolotas and Y. Demiris are with the Personal Robotics Lab, Dept. of Electrical and Electronic Engineering, Imperial College London,  SW7 2BT, UK; Email: \{mark.zolotas12, y.demiris\}@imperial.ac.uk}%
\thanks{This research was supported in part by an EPSRC Doctoral Training Award to MZ, and a Royal Academy of Engineering Chair in Emerging Technologies to YD.}%
}

\makeatletter
\def\ps@IEEEtitlepagestyle{%
  \def\@oddfoot{\mycopyrightnotice}%
  \def\@oddhead{\hbox{}\@IEEEheaderstyle\leftmark\hfil\thepage}\relax
  \def\@evenhead{\@IEEEheaderstyle\thepage\hfil\leftmark\hbox{}}\relax
  \def\@evenfoot{}%
}
\def\mycopyrightnotice{%
  \begin{minipage}{\textwidth}
  \centering \scriptsize
  Copyright~\copyright~20xx IEEE. Personal use of this material is permitted. Permission from IEEE must be obtained for all other uses, in any current or future media, including\\reprinting/republishing this material for advertising or promotional purposes, creating new collective works, for resale or redistribution to servers or lists, or reuse of any copyrighted component of this work in other works by sending a request to pubs-permissions@ieee.org.
  \end{minipage}
}
\makeatother

\begin{document}


\maketitle

\begin{abstract}
Equipping robots with the ability to infer human intent is a vital precondition for effective collaboration. Most computational approaches towards this objective derive a probability distribution of ``intent'' conditioned on the robot's perceived state. However, these approaches typically assume task-specific labels of human intent are known \textit{a priori}. To overcome this constraint, we propose the Disentangled Sequence Clustering Variational Autoencoder (\discvae{}), a clustering framework capable of learning such a distribution of intent in an \textit{unsupervised} manner. The proposed framework leverages recent advances in unsupervised learning to disentangle latent representations of sequence data, separating time-varying local features from time-invariant global attributes. As a novel extension, the \discvae{} also infers a discrete variable to form a latent mixture model and thus enable clustering over these global sequence concepts,~\eg\ high-level intentions. We evaluate the \discvae{} on a real-world human-robot interaction dataset collected using a robotic wheelchair. Our findings reveal that the inferred discrete variable coincides with human intent, holding promise for collaborative settings, such as shared control.
\end{abstract}
\section{Introduction}
\label{sec:intro}

Humans are remarkably proficient at inferring the implicit intentions of others from their overt behaviour. Consequently, humans are adept at planning their actions when collaborating together. Intention inference may therefore prove equally imperative in creating fluid and effective human-robot collaborations. Robots endowed with this ability have been extensively explored~\cite{Demiris2007,Losey2018,Jain2019}, yet their integration into real-world settings remains an open research problem.

One major impediment to real-world instances of robots performing human intention inference is the assumption that a known representation of intent exists. For example, most methods in collaborative robotics assume a discrete set of task goals is known \textit{a priori}. Under this assumption, the robot can infer a distribution of human intent by applying Bayesian reasoning over the entire goal space~\cite{Javdani2015,Jain2019}. Whilst such a distribution offers a versatile and practical representation of intent, the need for predefined labels is not always feasible unless restricted to a specific task scope. 

\begin{figure}[t]
  \centering
  \includegraphics[width=0.76\columnwidth]{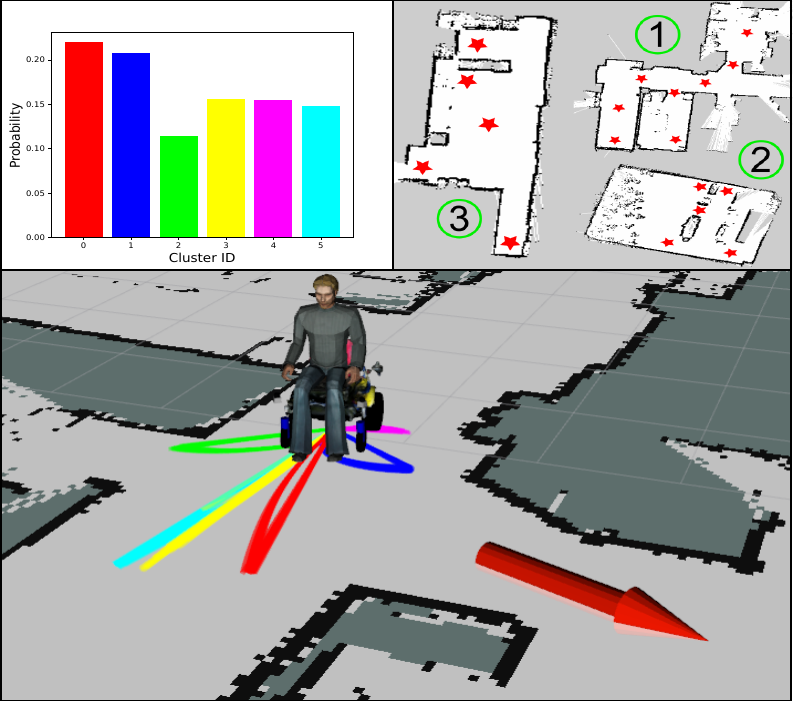}
  \caption{Overview visualisation of the intention inference experiment on a robotic wheelchair. \textbf{Bottom:} Recorded output of an \textit{actual} human subject navigating towards a goal (red arrow). \textbf{Top Right:} Maps of the three experiment settings, with red stars denoting target locations. \textbf{Top Left:} Probability histogram of the categorical variable modelling ``intentions'' at this particular snapshot of the data for $K\,{=}\,6$ clusters. The bars are coloured to align with the wheelchair trajectories generated by sampling from the corresponding clusters. Multiple diverse trajectories can be sampled from the same cluster and each trajectory's length is dependent on the velocity commands drawn from the generative model. Figure best viewed in colour.}
  \label{fig:exp_overview}
  \vspace{-3.4mm}
\end{figure}

Another fundamental challenge is that many diverse actions often fulfil the same intention. A popular class of probabilistic algorithms for overcoming this challenge are generative models, which derive a distribution of observations by introducing latent random variables to capture any hidden underlying structure. Within the confines of intention inference, the modelled latent space is then presumed to represent all possible causal relations between intentions and observed human behaviour~\cite{Wang2013,Tanwani2017,Hu2019}. The advent of deep generative models, such as Variational Autoencoders (VAEs)~\cite{Kingma2013,Rezende2014}, has also enabled efficient inference of this latent space from abundant sources of high-dimensional data.

Inspired by the prospects of not only extracting hidden ``intent'' variables but also interpreting their meaning, we frame the intention inference problem as a process of \textit{disentangling} the latent space. Disentanglement is a core research thrust in representation learning that refers to the recovery of abstract concepts from independent factors of variation assumed to be responsible for generating the observed data~\cite{Chen2018,Dupont2018,Locatello2019}.
The interpretable structure of these independent factors is exceedingly desirable for human-in-the-loop scenarios~\cite{Fortuin2018}, like robotic wheelchair assistance, however few applications have transferred over to the robotics domain~\cite{Hu2019}.

We strive to bridge this gap by proposing an \textit{unsupervised} disentanglement framework suitable for human intention inference. Capitalising on prior disentanglement techniques, we learn a latent representation of sequence observations that divides into a local (time-varying) and global (time-preserving) part~\cite{Yingzhen2018,Hsu2018}. Our proposed variant simultaneously infers a categorical variable to construct a mixture model and thereby form clusters in the \textit{global} latent space. In the scope of intention inference, we view the \textit{continuous} local variable as representative of desirable low-level trajectories, whilst the \textit{discrete} counterpart signifies high-level intentions. To summarise, this paper's contributions are:
\begin{itemize}
    \item A framework for clustering disentangled representations of sequences, coined as the \textit{Disentangled Sequence Clustering Variational Autoencoder (DiSCVAE)};
    \item Findings from a robotic wheelchair experiment (see Fig.~\ref{fig:exp_overview}) that demonstrate how clusters learnt without explicit supervision can be interpreted as user-intended navigation behaviours, or strongly correlated with ``labels'' of such intent in a semi-supervised context.
\end{itemize}

\section{Preliminaries}
\label{sec:preliminaries}



\begin{figure}[t]
    \centering
    \subfloat[VAE]{\includegraphics[width=0.22\columnwidth]{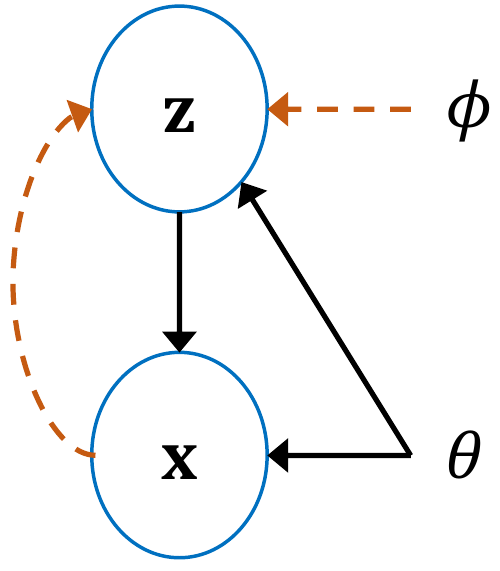}
    \label{fig:vae}}
    \hfil
    \subfloat[VRNN]{\includegraphics[width=0.28\columnwidth]{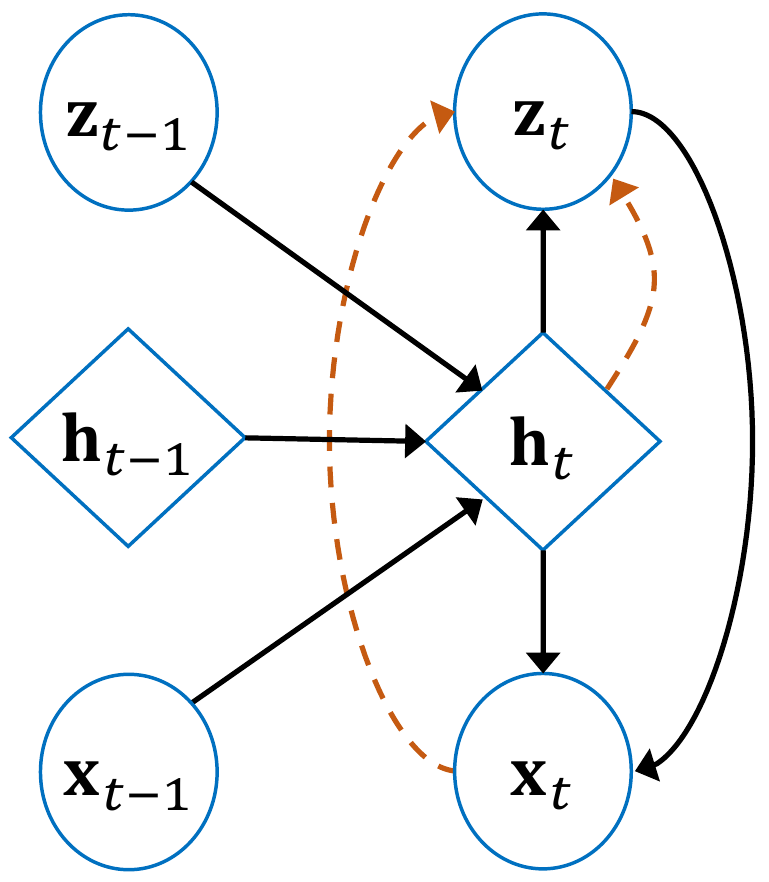}
    \label{fig:vrnn}}
    \hfil
    \subfloat[GMVAE]{\includegraphics[width=0.27\columnwidth]{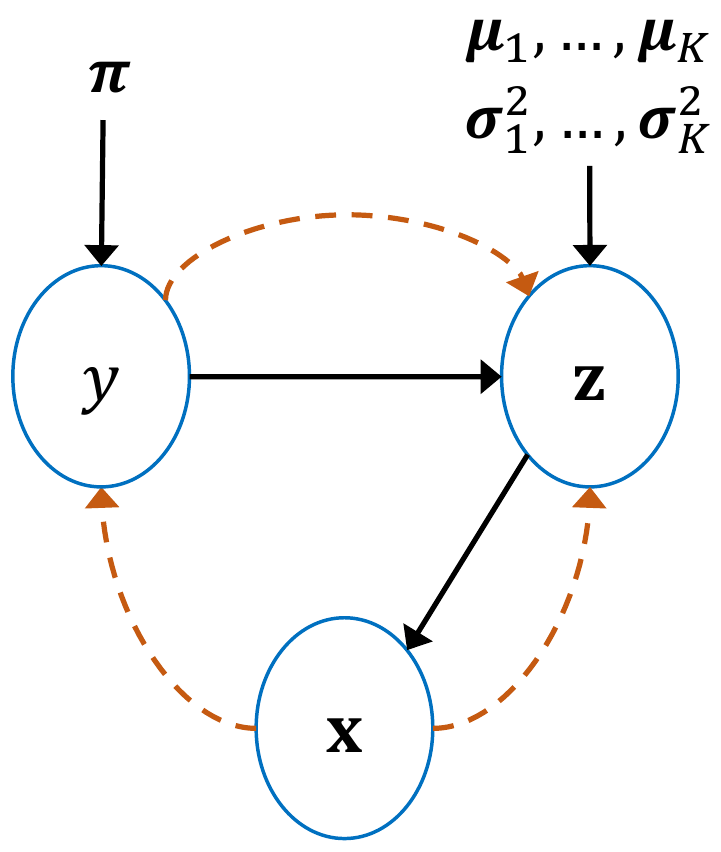}
    \label{fig:gmvae}}
    \caption{Deep generative models for: (a) variational inference~\cite{Kingma2013,Rezende2014}; (b) a sequential VAE that conditions on the deterministic hidden states of an RNN at each timestep (VRNN~\cite{Chung2015}); (c) a VAE with a Gaussian mixture prior (GMVAE). Dashed lines denote inference and bold lines indicate generation.} 
    \label{fig:preliminaries}
    \vspace{-2.4mm}
\end{figure}

Before defining the \discvae{}, we describe supporting material from representation learning, starting with the VAE displayed in Fig.~\ref{fig:vae}. The VAE is a deep generative model consisting of a generative and recognition network. These networks are jointly trained by applying the reparameterisation trick~\cite{Kingma2013,Rezende2014} and maximising the evidence lower bound (ELBO) $\mathcal{L}_{\theta,\phi} (\mathbf{x})$ on the marginal log-likelihood:
\begin{align}
    \log p_{\theta}(\mathbf{x}) & \geq \mathcal{L}_{\theta,\phi} (\mathbf{x}) \label{eq:vae_elbo}
    \\
    & \equiv \E_{q_{\phi}(\mathbf{z} | \mathbf{x})} \! \big[\log p_{\theta}(\mathbf{x} | \mathbf{z})\big] - \text{KL}\big({q_{\phi}(\mathbf{z} | \mathbf{x}) \,||\, p_{\theta}(\mathbf{z})}\big), \nonumber 
\end{align}
where the first term is the reconstruction error of reproducing observations $\mathbf{x}$, and the second KL divergence term is a regulariser that encourages the variational posterior $q_{\phi}(\mathbf{z}\,|\,\mathbf{x})$ to be close to the prior $p_{\theta}(\mathbf{z})$.
For notational convenience, parameters $\phi$ and $\theta$ will be omitted hereafter.


Deep generative models can also be parameterised by Recurrent Neural Networks (RNNs) to represent temporal data under the VAE learning principle. A notable example is the Variational RNN (VRNN)~\cite{Chung2015} shown in Fig.~\ref{fig:vrnn}, which conditions on latent variables and observations from previous timesteps via its deterministic hidden state, $\mathbf{h}_{t}(\mathbf{x}_{t-1},\mathbf{z}_{t-1},\mathbf{h}_{t-1})$, leading to the joint distribution:
\begin{align}
    p(\mathbf{x}_{\leq T}, \mathbf{z}_{\leq T}) & = \prod_{t=1}^T p(\mathbf{x}_t\,|\,\mathbf{z}_{\leq t},\mathbf{x}_{<t})p(\mathbf{z}_t\,|\,\mathbf{x}_{<t},\mathbf{z}_{<t}) \label{eq:vrnn_gen}
    \\
    & = \prod_{t=1}^T p(\mathbf{x}_t\,|\,\mathbf{z}_{t},\mathbf{h}_{t})p(\mathbf{z}_t\,|\,\mathbf{h}_{t}), \nonumber
\end{align}
where the true posterior is conditioned on information pertaining to previous observations $\mathbf{x}_{<t}$ and latent states $\mathbf{z}_{<t}$, hence accounting for temporal dependencies. The VRNN state $\mathbf{h}_{t}$ is also shared with the inference procedure to yield the variational posterior distribution:
\begin{equation} \label{eq:vrnn_inf}
    q(\mathbf{z}_{\leq T}\,|\,\mathbf{x}_{\leq T}) = \prod_{t=1}^T q(\mathbf{z}_t|\mathbf{x}_{\leq t},\mathbf{z}_{<t}) = \prod_{t=1}^T q(\mathbf{z}_t|\mathbf{x}_{t},\mathbf{h}_{t}).
\end{equation}

The \discvae{} developed in the following section elects an approach akin to the VRNN, where latent variables are injected into the forward autoregressive dynamics.
\section{Disentangled Sequence Clustering Variational Autoencoder}
\label{sec:discvae}

In this section, we introduce the \textit{Disentangled Sequence Clustering VAE (\discvae{})}\footnote{Code available at: \href{https://github.com/mazrk7/discvae}{\url{https://github.com/mazrk7/discvae}}}, a framework suited for human intention inference. Clustering is initially presented as a Gaussian mixture adaptation of the VAE prior. The complete \discvae{} is then specified by combining this adaptation with a sequential model that disentangles latent variables. Finally, we relate back to the intention inference domain.


\subsection{Clustering with Variational Autoencoders}
\label{sec:discvae:clustering}

A crucial aspect of generative models is choosing a prior capable of fostering structure or clusters in the data. Previous research has tackled clustering with VAEs by segmenting the latent space into distinct classes using a Gaussian mixture prior,~\ie\ a GMVAE~\cite{Dilokthanakul2016,Jiang2017}. 

Our approach is similar to earlier GMVAEs, except for two modifications. First, we leverage the categorical reparameterisation trick to obtain differentiable samples of discrete variables~\cite{Maddison2016,Jang2016}. Second, we alter the ELBO to mitigate the precarious issues of posterior collapse and cluster degeneracy~\cite{Hsu2018}. Posterior collapse refers to latent variables being ignored or overpowered by highly expressive decoders during training, such that the posterior mimics the prior. Whilst cluster degeneracy is when multiple modes of the prior have collapsed into one~\cite{Dilokthanakul2016}.

The GMVAE outlined below is the foundation for how the \discvae{} uncovers $K$ clusters (see Fig.~\ref{fig:gmvae}). Assuming observations $\mathbf{x}$ are generated according to some stochastic process with discrete latent variable $y$ and continuous latent variable $\mathbf{z}$, then the joint probability can be written as:
\begin{align}
    p(\mathbf{x},\mathbf{z},y) & = p(\mathbf{x}\,|\,\mathbf{z})p(\mathbf{z}\,|\,y)p(y) \label{eq:gmvae_gen} \\
    y & \sim \text{Cat}(\bm{\pi}) \nonumber \\
    \mathbf{z} & \sim \mathcal{N}\big(\bm{\mu}_{z}(y),\text{diag}(\bm{\sigma}^2_{z}(y))\big) \nonumber \\
    \mathbf{x} & \sim \mathcal{N}\big(\bm{\mu}_{x}(\mathbf{z}),\bm{I}\big) \; \text{or} \; \mathcal{B}\big(\bm{\mu}_{x}(\mathbf{z})\big), \nonumber
\end{align}
where functions $\bm{\mu}_{z}$, $\bm{\sigma}^2_{z}$ and $\bm{\mu}_{x}$ are neural networks whose outputs parameterise the distributions of $\mathbf{z}$ and $\mathbf{x}$. The generative process involves three steps: (1) sampling $y$ from a categorical distribution parameterised by probability vector $\bm{\pi}$ with $\pi_{k}$ set to $K^{-1}$; (2) sampling $\mathbf{z}$ from the marginal prior $p(\mathbf{z}\,|\,y)$, resulting in a Gaussian mixture with a diagonal covariance matrix and uniform mixture weights; and (3) generating data $\mathbf{x}$ from a likelihood function $p(\mathbf{x}\,|\,\mathbf{z})$.

A variational distribution $q(\mathbf{z},y\,|\,\mathbf{x})$ for the true posterior can then be introduced in its factorised form as:
\begin{equation} \label{eq:gmvae_inf}
    q(\mathbf{z},y\,|\,\mathbf{x}) = q(\mathbf{z}\,|\,\mathbf{x},y)q(y\,|\,\mathbf{x}),
\end{equation}
where both the multivariate Gaussian $q(\mathbf{z}\,|\,\mathbf{x},y)$ and categorical $q(y\,|\,\mathbf{x})$ are also parameterised by neural networks, with respective parameters, $\phi_z$ and $\phi_y$, omitted from notation.

Provided with these inference $q(.)$ and generative $p(.)$ networks, the ELBO for this clustering model becomes:
\begin{align}
    \mathcal{L}(\mathbf{x}) & = \E_{q(\mathbf{z}, y \,|\, \mathbf{x})} \bigg[\log \frac{p(\mathbf{x}, \mathbf{z}, y)}{q(\mathbf{z}, y \,|\, \mathbf{x})} \bigg] \label{eq:gmvae_elbo}
    \\
    & = \E_{q(\mathbf{z}, y \,|\, \mathbf{x})}\big[\log p(\mathbf{x} \,|\, \mathbf{z})\big] \nonumber
    \\ 
    & \quad -\> \E_{q(y \,|\, \mathbf{x})}\big[\text{KL}\big({q(\mathbf{z} \,|\, \mathbf{x}, y) \,||\, p(\mathbf{z} \,|\, y)}\big)\big] \nonumber
    \\
    & \quad -\> \text{KL}\big({q(y \,|\, \mathbf{x}) \,||\, p(y)}\big), \nonumber
\end{align}
where the first term is reconstruction loss of data $\mathbf{x}$, and the latter two terms push the variational posteriors close to their corresponding priors. As the standard reparameterisation trick is intractable for non-differentiable discrete samples, we employ a continuous relaxation of $q(y\,|\,\mathbf{x})$~\cite{Maddison2016,Jang2016} that removes the need to marginalise over all $K$ class values.

Optimising GMVAEs with powerful decoders is prone to cluster degeneracy due to the over-regularisation effect of the KL term on $y$ opting for a uniform posterior~\cite{Dilokthanakul2016}. As KL divergence is a known upper bound on mutual information between a latent variable and data during training~\cite{Dupont2018,Chen2018}, we instead penalise mutual information in Eq.~\ref{eq:gmvae_elbo} by replacing $\text{KL}\big(q(y \,|\, \mathbf{x}) \,||\, p(y)\big)$ with entropy $\mathcal{H}\big(q(y\,|\,\mathbf{x})\big)$ given uniform $p(y)$. We found this modification to be empirically effective at preventing mode collapse and it may even improve the other key trait of the \discvae{}: disentanglement~\cite{Dupont2018}.


\subsection{Model Specification}
\label{sec:discvae:model}

\begin{figure}
  \centering
  \includegraphics[width=0.95\columnwidth]{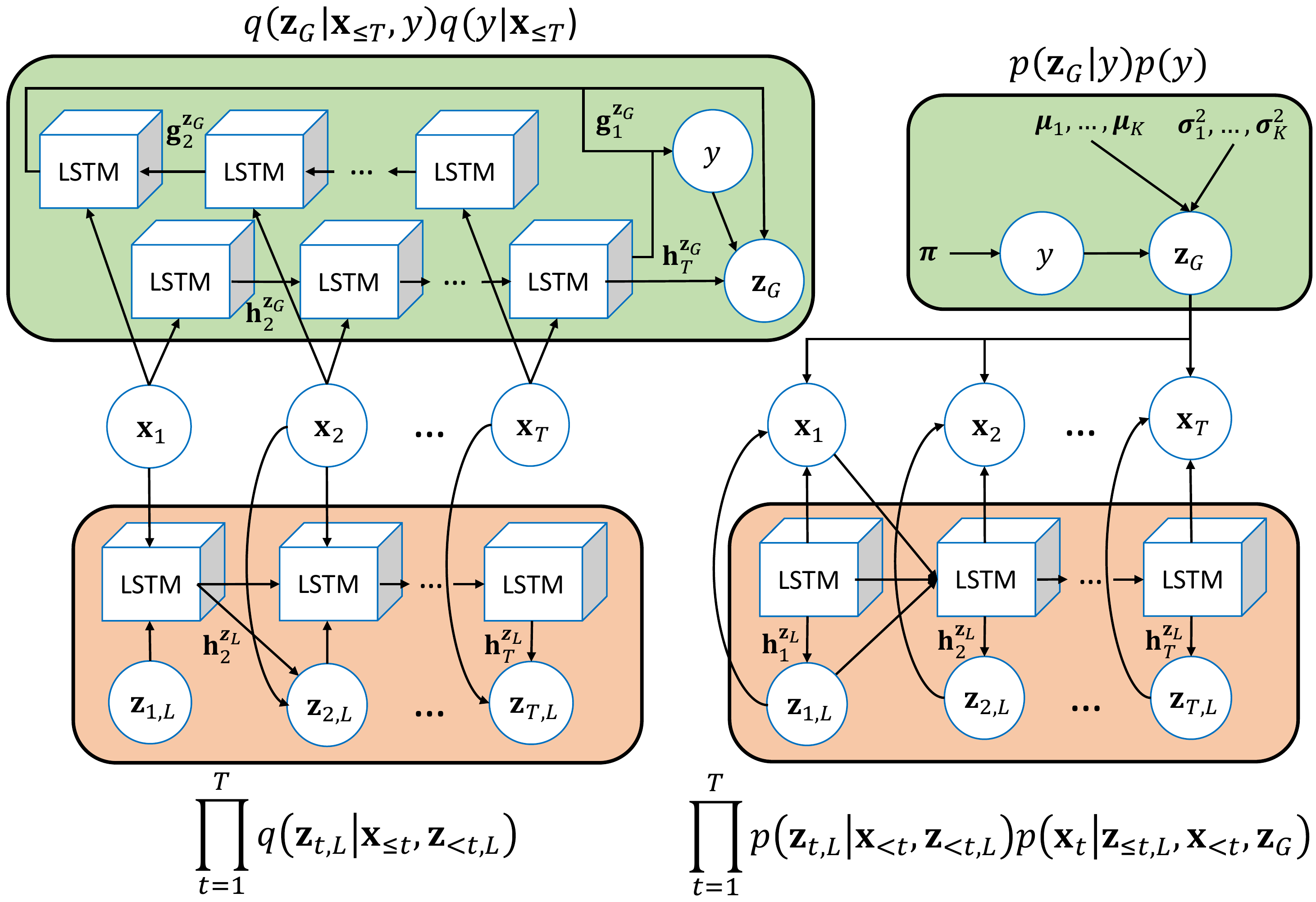}
  \caption{Computation graph of the inference $q(.)$ and generative $p(.)$ networks. \textcolor{green(ryb)}{Green} blocks contain global variables $y$ and $\mathbf{z}_G$, with a bidirectional LSTM conditioning over input sequence $\mathbf{x}_{\leq T}$. Forward $\mathbf{h}_t^{\mathbf{z}_G}$ and backward $\mathbf{g}_t^{\mathbf{z}_G}$ states then compute the $q(.)$ distribution parameters. \textcolor{orange(ryb)}{Orange} blocks encompass the local sequence variable $\mathbf{z}_{t,L}$, where an LSTM's states $\mathbf{h}_t^{\mathbf{z}_{L}}$ are combined at each timestep with current inputs $\mathbf{x}_t$ to infer $\mathbf{z}_{t,L}$. Generating $\mathbf{x}_t$ requires \textit{both} $\mathbf{z}_G$ and $\mathbf{z}_{t,L}$. Figure best viewed in colour.}
  \label{fig:discvae}
  \vspace{-3.5mm}
\end{figure}

Having established how to categorise the VAE latent space learnt over static data, we now derive the \discvae{} (see Fig.~\ref{fig:discvae}) as a sequential extension that automatically clusters and \textit{disentangles} representations. Disentanglement amongst sequential VAEs commonly partitions latent representations into \textit{time-invariant} and \textit{time-dependent} subsets~\cite{Yingzhen2018,Hsu2018}. Similarly, we express our disentangled representation of some input sequence $\mathbf{x}_{\leq T}$ at timestep $t$ as $\mathbf{z}_t\,{=}\,[\mathbf{z}_G, \mathbf{z}_{t,L}]$, where $\mathbf{z}_G$ and $\mathbf{z}_{t,L}$ encode \textit{global} and \textit{local} features.


The novelty of our approach lies in how we solely cluster the global variable $\mathbf{z}_G$ extracted from sequences. Related temporal clustering models have either mapped the entire sequence $\mathbf{x}_{\leq T}$ to a discrete latent manifold~\cite{Fortuin2018} or inferred a categorical factor of variation to cluster over an \textit{entangled} continuous latent representation~\cite{Hsu2018}. Whereas the \discvae{} clusters high-level attributes $\mathbf{z}_G$ in isolation from lower-level dynamics $\mathbf{z}_{t,L}$. Furthermore, this proposed formulation plays an important role in our interpretation of intention inference, as is made apparent in Section~\ref{sec:discvae:intent}.

Using the clustering scheme described in Section~\ref{sec:discvae:clustering}, we define the generative model $p(\mathbf{x}_{\leq T}, \mathbf{z}_{\leq T,L}, \mathbf{z}_G, y)$ as:
\begin{equation} \label{eq:discvae_gen}
 p(\mathbf{z}_G\,|\,y) p(y) \prod_{t=1}^T p(\mathbf{x}_t\,|\,\mathbf{z}_{t,L},\mathbf{z}_G,\mathbf{h}^{\mathbf{z}_{L}}_{t})p(\mathbf{z}_{t,L}\,|\,\mathbf{h}^{\mathbf{z}_{L}}_{t}).
\end{equation}
The mixture prior $p(\mathbf{z}_G\,|\,y)$ encourages mixture components (indexed by $y$) to emerge in the latent space of variable $\mathbf{z}_G$. Akin to a VRNN~\cite{Chung2015}, the posterior of $\mathbf{z}_{t,L}$ is parameterised by deterministic state $\mathbf{h}^{\mathbf{z}_{L}}_{t}$. We also highlight the dependency on both $\mathbf{z}_{t,L}$ and $\mathbf{z}_G$ upon generating $\mathbf{x}_t$.

To perform posterior approximation, we adopt the variational distribution $q(\mathbf{z}_{\leq T,L},\mathbf{z}_G,y\,|\,\mathbf{x}_{\leq T})$ and factorise it as:
\begin{equation} \label{eq:discvae_inf}
 q(\mathbf{z}_G\,|\,\mathbf{x}_{\leq T}, y)q(y\,|\,\mathbf{x}_{\leq T})\prod_{t=1}^T q(\mathbf{z}_{t,L}\,|\,\mathbf{x}_{t},\mathbf{h}^{\mathbf{z}_{L}}_{t}),
\end{equation}
with a differentiable relaxation of categorical $y$ injected into the process when training~\cite{Maddison2016,Jang2016}. 

Under the VAE paradigm, the \discvae{} is trained by maximising the time-wise objective:
\begin{align}
    \mathcal{L}(\mathbf{x}_{\leq T}) & = \E_{q(\cdot)} \bigg[\log \frac{p(\mathbf{x}_{\leq T}, \mathbf{z}_{\leq T,L}, \mathbf{z}_G, y)}{q(\mathbf{z}_{\leq T,L},\mathbf{z}_G,y\,|\,\mathbf{x}_{\leq T})} \bigg] \label{eq:discvae_elbo}
    \\ & = \E_{q(\cdot)} \bigg[ \sum_{t=1}^T \bigg(\log p(\mathbf{x}_t\,|\,\mathbf{z}_{t,L},\mathbf{z}_G,\mathbf{h}^{\mathbf{z}_{L}}_{t}) \nonumber
    \\
    & \quad -\> \text{KL}\big(q(\mathbf{z}_{t,L}\,|\,\mathbf{x}_{t},\mathbf{h}^{\mathbf{z}_{L}}_{t}) \,||\, p(\mathbf{z}_{t,L}\,|\,\mathbf{h}^{\mathbf{z}_{L}}_{t})\big) \bigg) \nonumber
    \\ 
    & \quad -\> \text{KL}\big(q(\mathbf{z}_G\,|\,\mathbf{x}_{\leq T}, y) \,||\, p(\mathbf{z}_G\,|\,y) \big) \nonumber
    \\ 
    & \quad +\> \mathcal{H}\big(q(y\,|\,\mathbf{x}_{\leq T})\big) \bigg]. \nonumber
\end{align}
This summation of lower bounds across timesteps is decomposed into: (1) the expected log-likelihood of input sequences; (2) KL divergences for variables $\mathbf{z}_{t,L}$ and $\mathbf{z}_G$; and (3) entropy regularisation to alleviate mode collapse.


\subsection{Network Architecture}
\label{sec:discvae:arch}

The \discvae{} is graphically illustrated in Fig.~\ref{fig:discvae}. An RNN parameterises the posteriors over $\mathbf{z}_{t,L}$, with the hidden state $\mathbf{h}^{\mathbf{z}_{L}}_{t}$ allowing $\mathbf{x}_{<t}$ and $\mathbf{z}_{<t,L}$ to be indirectly conditioned on in Eqs.~\ref{eq:discvae_gen} and~\ref{eq:discvae_inf}. For time-invariant variables $y$ and $\mathbf{z}_G$, a bidirectional RNN extracts feature representations from the entire sequence $\mathbf{x}_{\leq T}$, analogous to prior architectures~\cite{Yingzhen2018}. Bidirectional forward $\mathbf{h}_t$ and reverse $\mathbf{g}_t$ states are computed by iterating through $\mathbf{x}_{\leq T}$ in both directions, before being merged by summation. RNNs have LSTM cells and multilayer perceptrons (MLPs) are dispersed throughout to output the mean and variance of Gaussian distributions. 


\subsection{Intention Inference}
\label{sec:discvae:intent}

\begin{algorithm}[t]
\caption{Sampling to produce diverse predictions of goal states from the inferred cluster $c$}
\label{alg:generative_procedure}
    \textbf{Input}: Observation sequence $\mathbf{x}_{\leq t}$; sample length $n$; \\
    \textbf{Initialise}: $\mathbf{h}_t \leftarrow \mathbf{0}$; $\mathbf{z}_{t, L} \leftarrow \mathbf{0}$;\\
    \textbf{Output}: Predicted states $\tilde{\mathbf{x}}_{t+1},\ldots,\tilde{\mathbf{x}}_{t+n}$
    
    \begin{algorithmic} 
    \STATE Feed prefix $\mathbf{x}_{\leq t}$ into inference model (Eq.~\ref{eq:discvae_inf})
    \STATE Assign to cluster $c$ (Eq.~\ref{eq:cluster_assignment})
    \STATE Draw fixed global sample from $p(\mathbf{z}_G\,|\,y=c)$
    \FOR{$i \in \{t+1, \ldots, t+n\}$}
    \STATE Update $\mathbf{h}_{i} \leftarrow \text{RNN}(\mathbf{z}_{i-1, L},\mathbf{x}_{i-1},\mathbf{h}_{i-1})$
    \STATE Sample local dynamics from $p(\mathbf{z}_{i,L} \,|\, \mathbf{h}_{i})$
    \STATE Predict $\tilde{\mathbf{x}}_i \sim p(\mathbf{x}_i \,|\, \mathbf{z}_{i,L},\mathbf{h}_{i}, \mathbf{z}_G)$
    \ENDFOR
    \end{algorithmic}
\end{algorithm}

Let us now recall the problem of intention inference. We first posit that the latent class attribute $y$ could model a $K$-dimensional repertoire of action plans when considering human interaction data for a specific task. From this perspective, intention inference is a matter of assigning clusters (or action plans) to observations $\mathbf{x}_{\leq T}$ of human behaviour and the environment (\eg\ joystick commands and sensor data). Human intent is thus computed as the most probable element of the component posterior:
\begin{equation} \label{eq:cluster_assignment}
   c = \argmax_k q(y_k\,|\,\mathbf{x}_{\leq T}),
\end{equation}
where $c$ is the assigned cluster identity,~\ie\ the inferred intention label. The goal associated with this cluster is then modelled by $\mathbf{z}_G$, and local variable $\mathbf{z}_{t,L}$ captures the various behaviours capable of accomplishing the inferred plan.

In the robotic wheelchair scenario, most related works on intention estimation represent user intent~\cite{Carlson2012,Poon2017} as a target wheelchair state $\tilde{\mathbf{x}}_T$. Bayesian reasoning over the entire observation sequence $\mathbf{x}_{\leq T}$ using an entangled latent variable can yield such a state~\cite{Wang2013,Tanwani2017,Jain2019}. In contrast, the \discvae{} employs a disentangled representation $\mathbf{z}_t\,{=}\,[\mathbf{z}_G, \mathbf{z}_{t,L}]$, where the goal state variable is explicitly separated from the user action and environment dynamics. The major benefit of this separation is controlled generation, where repeatedly sampling $\mathbf{z}_{t,L}$ can enable diversity in how trajectories $\tilde{\mathbf{x}}_t$ pan out according to the global plan. The procedure for inferring intention label $c$ amongst a collection of action plans and generating diverse trajectories is summarised in Algorithm~\ref{alg:generative_procedure}.

\section{Intention Inference on Robotic Wheelchairs}
\label{sec:wheelchair}

To validate the \discvae{} utility at intention inference, we consider a dataset of real users navigating a wheelchair. The objective here is to infer user-intended action plans from observations of their joystick commands and surroundings.


\subsection{Dataset}
\label{sec:wheelchair:dataset}

Eight healthy subjects (aged 25-33) with experience using a robotic wheelchair were recruited to navigate three mapped environments (top right of Fig.~\ref{fig:exp_overview}). 
Each subject was requested to manually control the wheelchair using its joystick and follow a random route designated by goal arrows appearing on a graphical interface, as in Fig.~\ref{fig:exp_overview}. 

Experiment data collected during trials were recorded at a rate of 10 Hz, with sequences of length $T\,{=}\,20$. This sequence length $T$ is inspired by related work on estimating the short-term intentions of robotic wheelchair operators~\cite{Poon2017}. Every sequence was composed of user joystick commands $\mathbf{a}_t \,{\in}\, \mathbb{R}^2$ (linear and angular velocities), as well as rangefinder readings $\mathbf{l}_t \,{\in}\, \mathbb{R}^{360}$ ($1^{\circ}$ angular resolution), with both synchronised to the elected system frequency. The resulting dataset amounted to a total of 8883 sequences.

To assess the generalisability of our intention inference framework, we segregate the dataset based on the experiment environment. As a result, trials that took place in Map 3 are excluded from the training and validation sets, leaving splits of 5881/1580/1422 for training/testing/validation. Dividing the dataset in this way allows us to investigate performance under variations in task context, verifying whether the \discvae{} can elucidate human intent irrespective of such change. 


\subsection{Labelling Routine}
\label{sec:wheelchair:label}

Even without access to predefined labels for manoeuvres made by subjects while pursuing task goals, we can appoint approximations of user ``intent'' to shed light on the analysis. As such, an automated labelling routine is devised below.

Each sequence is initially categorised as either ``narrow'' or ``wide'' depending on a measure of threat applied in obstacle avoidance for indoor navigation~\cite{Durham2008}:
\begin{equation} \label{eq:threat}
 s_t = \frac{1}{N}\sum_{i=1}^N\text{sat}_{[0,1]} \bigg( \frac{D_s + R - l^i_t}{D_s} \bigg),
\end{equation}
where the aggregate threat score $s_t$ at timestep $t$ for $N\,{=}\,360$ laser readings $l^i_t$ is a saturated function of these ranges, the robot's radius $R$ (0.5 m for the wheelchair), and a safe distance parameter $D_s$ (set to 0.8 m). In essence, this score reflects the danger of imminent obstacles and qualifies narrow sequences whenever it exceeds a certain threshold. 

Next, we discern the intended navigation manoeuvres of participants from the wheelchair's odometry. After empirically testing various thresholds for translational and angular velocities, we determined six manoeuvres: in-place rotations (left/right), forward and reverse motion, as well as forward turns (left/right). This results in 12 classes that account for the influence of both the environment and user actions. Referring to Fig.~\ref{fig:exp_overview}, the majority class across Maps 1 \& 2 is the wide in-place rotation (left \textit{and} right), whilst for Map 3 it is the narrow reverse. This switch in label frequency highlights the task diversity caused by different maps.


\subsection{Implementation}
\label{sec:wheelchair:impl}

\begin{figure}[t]
  \centering
  \includegraphics[width=\columnwidth]{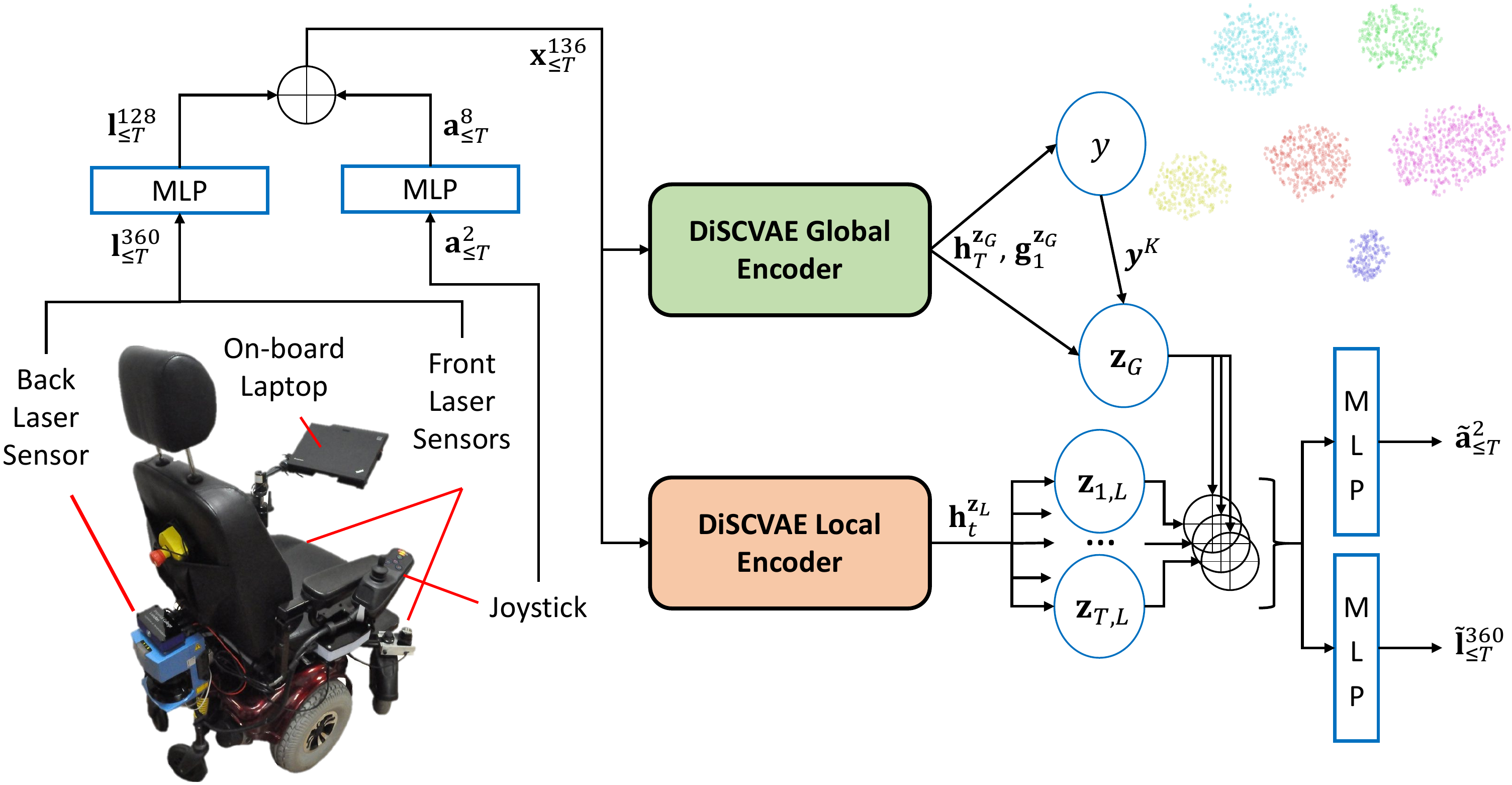}
  \caption{Architecture for the robotic wheelchair experiment. Joystick and laser data are fed into separate MLPs to produce a concatenated sequence, $\mathbf{x}_t \,{\in}\, \mathbb{R}^{136}$, which feeds into the \discvae{} encoder (Fig.~\ref{fig:discvae}). Forward and backward states, $\mathbf{h}^{\mathbf{z}_G}_{T}$ and $\mathbf{g}^{\mathbf{z}_G}_{1}$, allow inference of $\mathbf{z}_G$, whilst $\mathbf{z}_{t,L}$ instead conditions on hidden state $\mathbf{h}^{\mathbf{z}_L}_{t}$. These latent variables are then passed onto MLPs that decode the joystick commands $\tilde{\mathbf{a}}_t$ and range values $\tilde{\mathbf{l}}_t$.}
  \label{fig:wheelchair_arch}
  \vspace{-1.8mm}
\end{figure}

The robotic wheelchair has an on-board computer and three laser sensors, two at the front and one at the back for a full $360^{\circ}$ field of view. For readers interested in the robotic platform, please refer to our earlier work~\cite{Zolotas2019}.

Fig.~\ref{fig:wheelchair_arch} portrays the network architecture for this experiment. 
Before entering the network, input sequences are normalised per modality using the mean and standard deviation of the training set. To process the two input modalities, laser readings $\mathbf{l}_{\leq T}$ and control commands $\mathbf{a}_{\leq T}$ are first passed through separate MLPs. The derived code vectors are then concatenated $\mathbf{x}_{\leq T}$ and fed into the \discvae{} encoder to infer latent variables $\mathbf{z}_{G}$ and $\mathbf{z}_{\leq T,L}$. Two individual decoders are conditioned on these variables to reconstruct the original input sequences. Both sensory modalities are modelled as Gaussian variables with fixed variance. No odometry information was supplied at any point to this network.


\subsection{Evaluation Protocol \& Model Selection}
\label{sec:wheelchair:hyper}

The evaluation protocol for this experiment is as follows. Although labelled data are unavailable in most practical settings, including ours, we are still interested in digesting the prospects of the \discvae{} for downstream tasks, such as semi-supervised classification. Accordingly, we train a k-nearest neighbour (KNN) classifier over the learnt latent representation, $\mathbf{z}_{G}$, and judge intention estimation performance using two pervasive classification metrics: accuracy and the F1-score. Another typical measure in the field is mean squared error (MSE)~\cite{Tanwani2017}, hence we compare trajectory predictions of user actions $\tilde{\mathbf{a}}_t$ and laser readings $\tilde{\mathbf{l}}_t$ for 10 forward sampled states against ``ground truth'' future states.

Using this protocol, model selection was conducted on the holdout validation set. A grid search over the network hyperparameters found 512 hidden units to be suitable for the single-layer MLPs (ReLU activations) and bidirectional LSTM states. More layers and hidden units garnered no improvements in accuracy and overall MSE. However, 128 units was chosen for the shared $\mathbf{h}_t^{\mathbf{z}_{L}}$ state, as higher values had the trade-off of enhancing MSE but worsening accuracy, and so we opted for better classification. Table~\ref{tab:intent:hyper_select} also reports on the dimensionality effects of global $\mathbf{z}_G$ and local $\mathbf{z}_{t,L}$ for a fixed model setting. The most noteworthy pattern observed is the steep fall in accuracy when $\text{dim}(\mathbf{z}_{t,L})\,{>}\,16$. Given that smaller latent spaces raised MSE, a balanced dimensionality of 16 was configured for local and global features.

 \begin{table}[t]
\caption{Hyperparameter Selection on Validation Set}
\label{tab:intent:hyper_select}
\centering
\footnotesize
\setlength\tabcolsep{4pt} 
\begin{tabular}{l|ccc|ccc|ccc} 
\toprule
$\text{dim}(\mathbf{z}_G)$ & \multicolumn{3}{c}{16} & \multicolumn{3}{c}{32} & \multicolumn{3}{c}{64} \\
$\text{dim}(\mathbf{z}_{t,L})$ & 16 & 32 & 64 & 16 & 32 & 64 & 16 & 32 & 64 \\
\midrule
Acc (\%) $\uparrow$ & \textbf{77.9} & 54.3 & 22.9 & 72.9 & 41.4 & 15 & 74.9 & 28.8 & 14.5 \\
MSE $\downarrow$ & 4.52 & 4.56 & \textbf{4.43} & 4.69 & 4.69 & 4.45 & 4.47 & 4.55 & 4.5 \\
\bottomrule
\end{tabular}
\vspace{-.7mm}
\end{table}

Another core design choice of the \discvae{} is to select the number of clusters $K$. Without access to ground truth labels, we rely on an unsupervised metric, known as Normalised Mutual Information (NMI), to assess clustering quality. The NMI score occupies the range $[0,1]$ and is thus unaffected by different $K$ clusterings. This metric has also been used amongst similar VAEs for discrete representation learning~\cite{Fortuin2018}. Table~\ref{tab:intent:k_select} provides NMI scores as $K$ varies, where $K\,{=}\,13$ was settled on due to its marginal superiority and resemblance to the class count from Section~\ref{sec:wheelchair:label}.

\begin{table}[t]
\caption{Normalised Mutual Information to determine $K$}
\label{tab:intent:k_select}
\centering
\footnotesize
\setlength\tabcolsep{4pt} 
\begin{tabular}{l|ccccccc}  
\toprule
No. Clusters $K$ & 4 & 6 & 10 & 13 & 16 & 25 & 36 \\
\midrule
NMI $\uparrow$ & 0.141 & 0.133 & 0.206 & \textbf{0.264} & 0.244 & 0.24 & 0.26 \\
\bottomrule
\end{tabular}
\vspace{-3.1mm}
\end{table}


\subsection{Experimental Results}
\label{sec:wheelchair:results}

Six methods are considered in this experiment, each imitating the same network structure as in Fig.~\ref{fig:wheelchair_arch}:
\begin{itemize}
    \item \textbf{HMM:} A ubiquitous baseline in the literature~\cite{Tanwani2017,Jain2019};
    \item \textbf{SeqSVM:} A sequential SVM baseline~\cite{Wang2013};
    \item \textbf{BiLSTM:} A bidirectional LSTM classifier, akin to~\cite{Nicolis2018};
    \item \textbf{VRNN:} An autoregressive VAE model~\cite{Chung2015};
    \item \textbf{DSeqVAE:} A disentangled sequential autoencoder~\cite{Yingzhen2018};
    \item \textbf{DiSCVAE:} The proposed model of Section~\ref{sec:discvae:model}.
\end{itemize}
The top three \textit{supervised} models learn mappings between the inputs and labels identified in Section~\ref{sec:wheelchair:label}, where baselines utilised the trained BiLSTM encoder for feature extraction. Meanwhile, the bottom three VAE-based methods optimise their respective ELBOs, with a KNN trained on learnt latent variables for a \textit{semi-supervised} approach. Hyperparameters are consistent across methods,~\eg\ equal dimensions for the static and global latent variables of the DSeqVAE and \discvae{}, respectively. The Adam optimiser~\cite{Kingma2014b} was used to train models with a batch size of 32 and initial learning rate of $10^{-3}$ that exponentially decayed by 0.5 every 10k steps. From the range $3{\times}10^{-3}$ to $10^{-4}$, this learning rate had the most stable and effective ELBO optimisation performance. All models were optimised for 10 runs at different random seeds with early stopping ($\sim$75 epochs for the \discvae{}).

\begin{figure}[t]
  \centering
  \includegraphics[width=0.66\columnwidth]{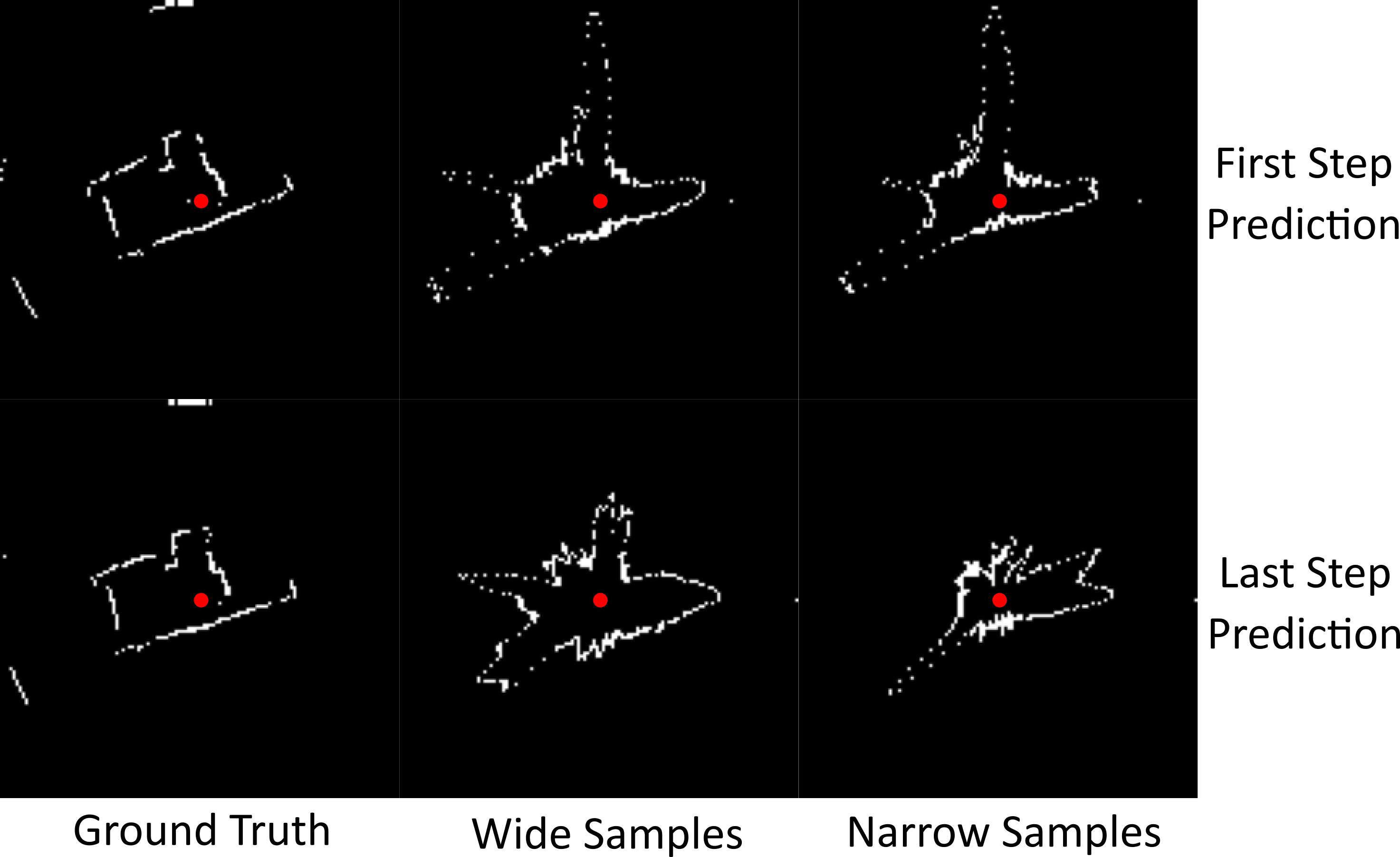}
  \caption{
  2D grids of predicted laser scans on the test set when sampling from ``wide'' and ``narrow'' type clusters. Wide samples create spacious proximity around the wheelchair (red dot), whilst narrow samples enclose space.}
  \label{fig:states}
  \vspace{-2mm}
\end{figure}

For qualitative analysis, a key asset of the \discvae{} is that sampling states from different clusters can exhibit visually diverse characteristics. Fig.~\ref{fig:exp_overview} portrays sampled trajectories from each mixture component during a subject's recorded interaction. There is clear variability in the trajectory outcomes predicted at this wheelchair configuration ($K\,{=}\,6$ to ease trajectory visualisation). The histogram over categorical $y$ (top left of Fig.~\ref{fig:exp_overview}) also indicates that the most probable trajectory aligns with the wheelchair user's current goal (red arrow),~\ie\ the correct ``intention''. As for generating future environment states, Fig.~\ref{fig:states} displays how samples from clusters manifest when categorised as either ``wide'' or ``narrow''.

Table~\ref{tab:wheelchair} contains quantitative results for this experiment. As anticipated, the highly variable nature of wheelchair control in an unconstrained navigation task makes classifying intent challenging. The baselines perform poorly and even the supervised BiLSTM obtains a classification accuracy of merely 56.3\% on the unseen test environment. Nevertheless, learning representations of user interaction data can reap benefits in intention inference, as performance is drastically improved by a KNN classifier trained over the latent spaces of the VAE-based methods. The \discvae{} acquires the best accuracy, F1-scores and MSE on joystick commands. The DSeqVAE instead attains the best error rates on forecasted laser readings at the expense of under-representing the relevant low-dimensional joystick signal. Cluster specialisation in the \discvae{} may explain the better $\tilde{\mathbf{a}}_{MSE}$.


\begin{table}[t]
    \caption{Performance on Test Set (10 random seeds)}
    \label{tab:wheelchair}
    \centering
    \footnotesize
    \setlength\tabcolsep{5pt} 
    \begin{tabular}{l|ccccc}  
    \toprule
    Model  & Acc (\%) $\uparrow$ & F1 $\uparrow$ & $\tilde{\mathbf{a}}_{MSE} \downarrow$ & $\tilde{\mathbf{l}}_{MSE} \downarrow$\\ 
    \midrule
    HMM & 12.3 $\pm$ 2.9 & 0.09 $\pm$ 0.02 & - & - \\
    SeqSVM & 48.3 $\pm$ 0.9 & 0.41 $\pm$ 0.01 & - & - \\
    BiLSTM & 56.3 $\pm$ 1.9 & 0.43 $\pm$ 0.02 & - & - \\
    VRNN & 65.1 $\pm$ 2.8 & 0.58 $\pm$ 0.04 & 0.15 $\pm$ 0.02 & 2.8 $\pm$ 0.04 \\
    DSeqVAE & 73.2 $\pm$ 2.0 & 0.65 $\pm$ 0.02 & 0.26 $\pm$ 0.02 & \textbf{2.14} $\pm$ \textbf{0.06} \\
    \discvae{} & \textbf{82.3} $\pm$ \textbf{1.8} & \textbf{0.78} $\pm$ \textbf{0.03} & \textbf{0.14} $\pm$ \textbf{0.01} & 2.7 $\pm$ 0.05 \\
    \bottomrule
    \end{tabular}
    \vspace{-4.7mm}
\end{table}


\subsection{Illuminating the Clusters}
\label{sec:wheelchair:clusters}

\begin{figure}[t]
  \centering
  \subfloat[Wheelchair Manoeuvres]{\includegraphics[width=0.46\columnwidth]{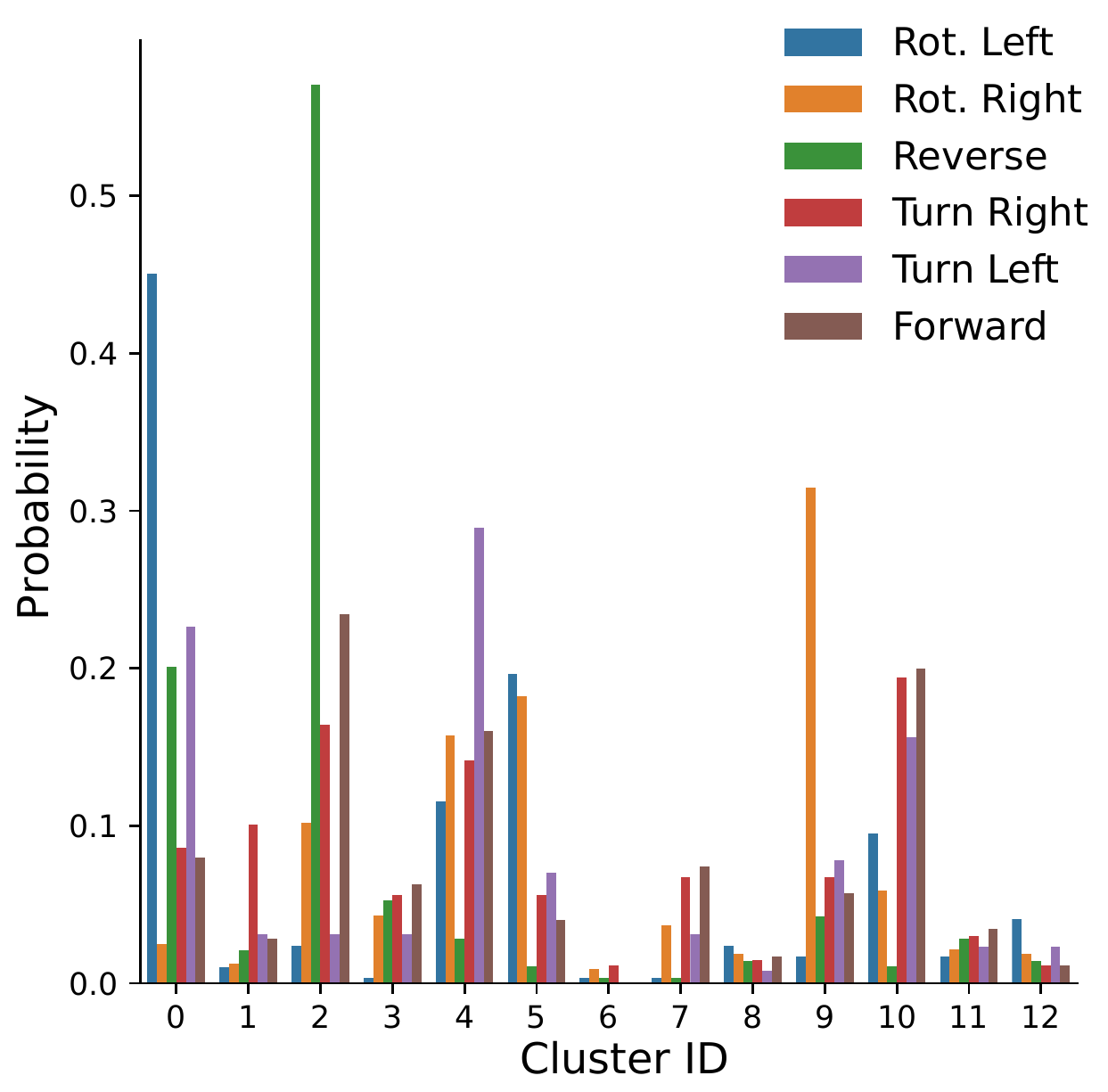}
  \label{fig:assign_action}}
  \hfil
  \subfloat[Spatial States]{\includegraphics[width=0.46\columnwidth]{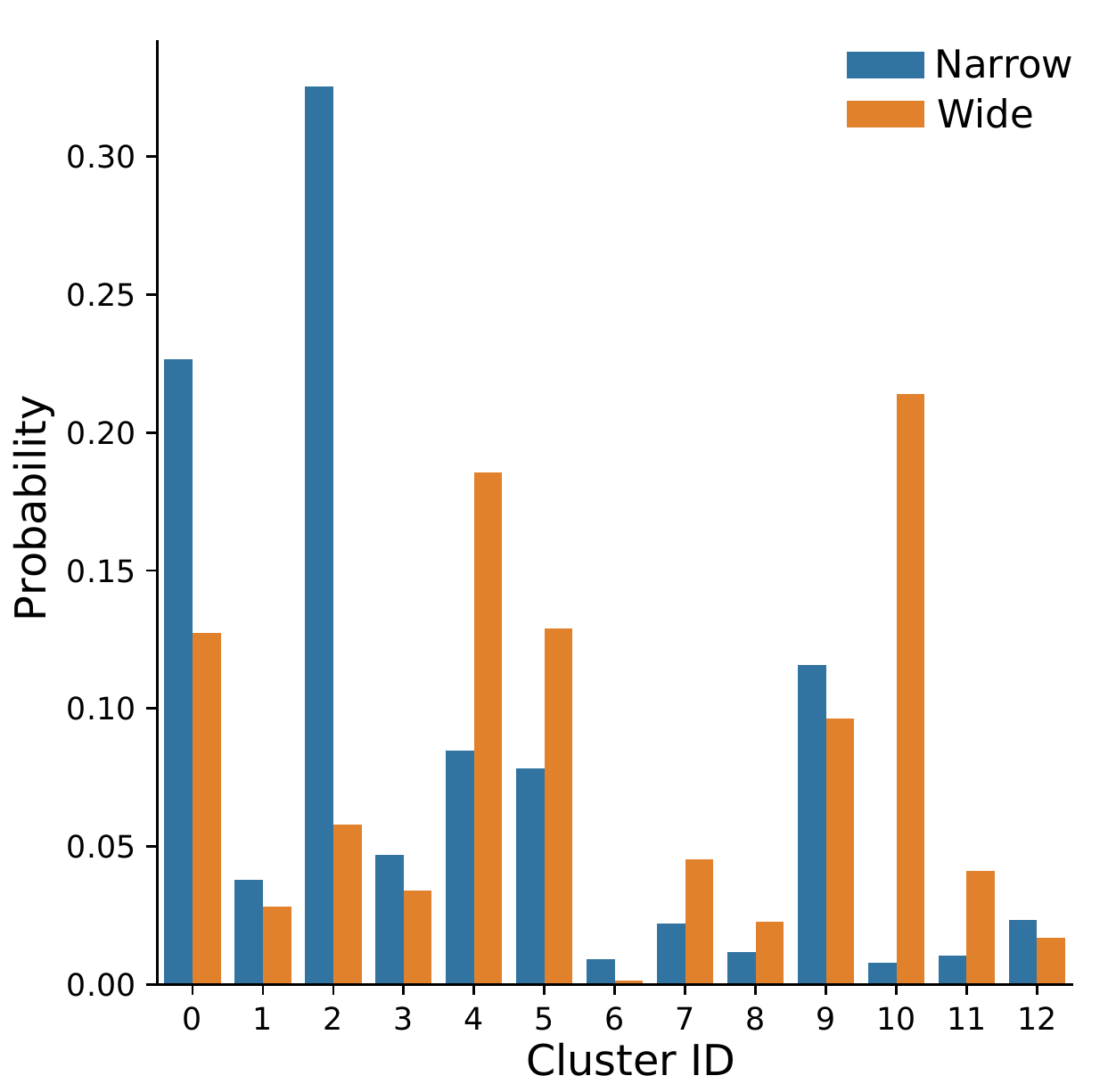}
  \label{fig:assign_state}}
  \caption{Assignment distribution of $y$ for $K\,{=}\,13$ with post-processed labels for \textbf{(a)} wheelchair manoeuvres and \textbf{(b)} perceived spatial context. The plot illuminates how various clusters are associated with user intent under different environmental conditions. For example, most backward motion and ``narrow'' state samples reside in cluster 2. Similar patterns are noticeable for in-place rotations (0 and 9) and ``wide'' forward motion (4 and 10).}
  \label{fig:assignment_y}
  \vspace{-2mm}
\end{figure}


Straying away from the purely discriminative task of classifying intent, we now use our framework to decipher the navigation behaviours, or ``global'' factors of variation, intended by users. In particular, we plot assignment distributions of $y$ on the test set examples to understand the underlying meaning of our clustered latent space. ``Local'' factors of variation in this application capture temporal dynamics in state,~\eg\ wheelchair velocities. 

Fig.~\ref{fig:assign_action} provides further clarity on how certain clusters have learnt independent wheelchair manoeuvres. For instance, cluster 2 is distinctly linked with the wheelchair's reverse motion. Likewise, clusters 0 and 9 pair with left and right in-place rotations. The spatial state assignments shown in Fig.~\ref{fig:assign_state} also delineate how these clusters are most often categorised as ``narrow'', which is to be expected of evasive actions taking place in cluttered spaces. On the contrary, predominantly forward-oriented manoeuvres fall into ``wide'' clusters (\eg\ 4 and 10). These findings suggest that wheelchair action plans have been aptly inferred.


\subsection{Prospects for Shared Control}
\label{sec:wheelchair:sc}

\begin{figure}[t]
  \centering
  \includegraphics[width=0.60\columnwidth]{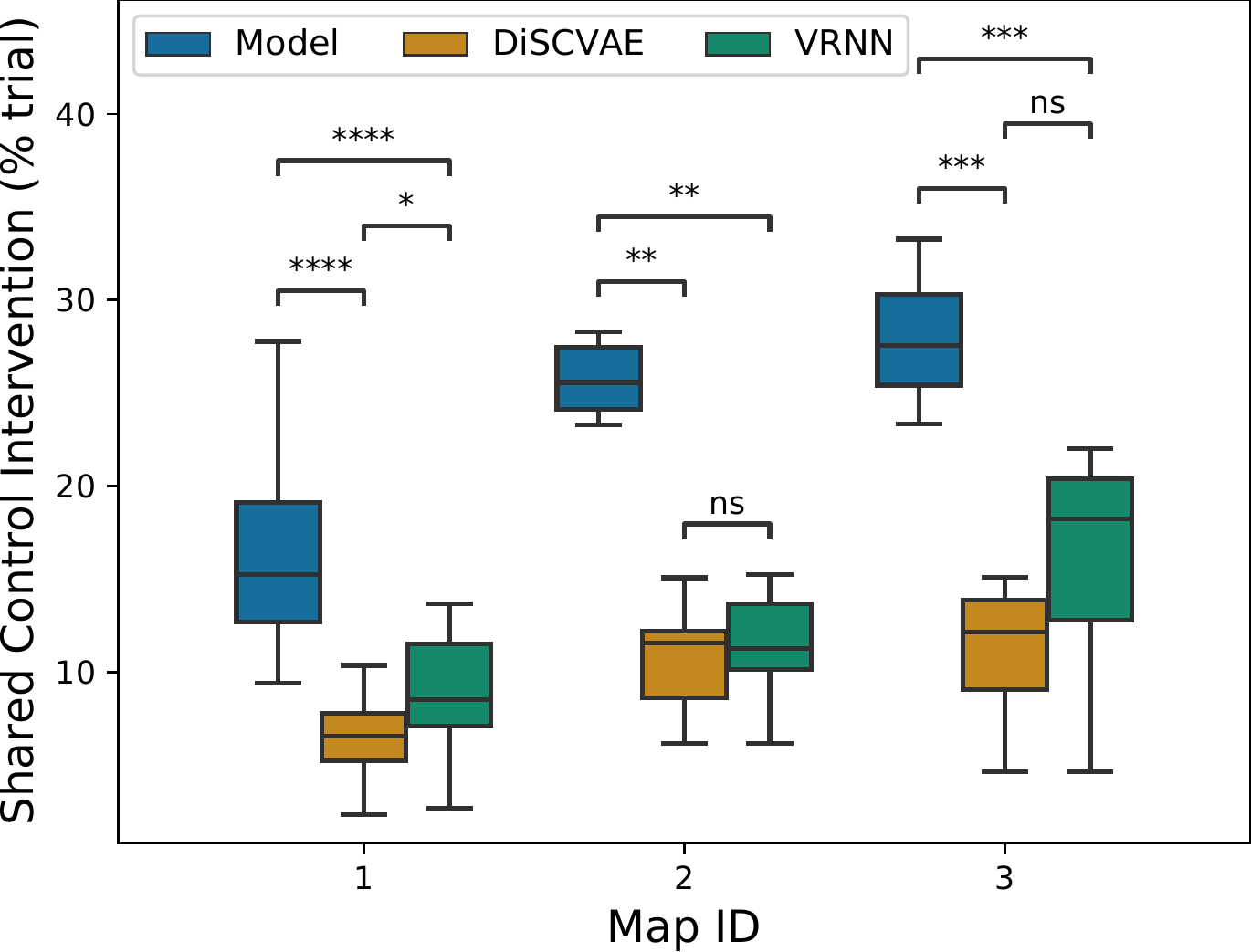}
  \caption{Percentage of trials per map where shared control would have wrongly intervened. The Model approach is significantly more likely to trigger incorrect assistance. Less variable VRNN and \discvae{} performance across maps also hints at better robustness to changes in task conditions.}
  \label{fig:sc_intervention}
  \vspace{-4.3mm}
\end{figure}

Lastly, we examine a shared control use-case, where intention inference plays a vital role~\cite{Demiris2007}. Shared control concerns the interaction between robots and humans when \textit{both} exert control over a system to accomplish a common goal~\cite{Losey2018}. Despite shared control being inactive for this experiment, we simulate its operation in post-processing to gauge success.

More precisely, we address the known issue in shared control of administering \textit{wrong} assistance whenever there is a misalignment between the robot's and user's internal models. To quantify this mismatch, we monitor the percentage of each navigation trial where a shared control methodology~\cite{Zolotas2019} would have intervened had it been operational. Given how the subjects are experienced, healthy individuals that incurred no wheelchair collisions, it is safe to assume they never required assistance. We compare wheelchair trajectories produced by the VRNN, \discvae{}, and a constant velocity ``Model'' using differential drive kinematics.

Fig.~\ref{fig:sc_intervention} offers results on shared control intervention rates. Performing the two-sided Mann-Whitney U test finds significantly better rates for the VRNN and \discvae{} over the Model across all maps ($p \! \leq \! 0.01$). Excluding Map 1 ($p  \! \leq  \! 0.05$), the positive trend in the \discvae{} surpassing the VRNN is not significant. Though the \discvae{} has the advantage of capturing uncertainty around its estimated intent via the categorical $y$,~\eg\ when a strict left-turn is hard to distinguish from a forward left-turn (blue and red in Fig.~\ref{fig:exp_overview}). This holds potential for shared control seeking to realign mismatched internal models by explaining to a user why the robot chose not to assist under uncertainty~\cite{Zolotas2019}.




\section{Discussion}
\label{sec:discuss}

There are a few notable limitations to this work. One is that learning disentangled representations is sensitive to hyperparameter tuning, as shown in Section~\ref{sec:wheelchair:hyper}. To aid with model selection and prevent posterior collapse, further investigation into different architectures and other information theoretic advances is thus necessary~\cite{Chen2018,Dupont2018}. Moreover, disentanglement and interpretability are difficult to define, often demanding access to labels for validation~\cite{Chen2018,Locatello2019}. Therefore, a study into whether users believe the \discvae{} representations of intent are ``interpretable'' or helpful for the wheelchair task is integral in claiming disentanglement.

In human-robot interaction tasks, intention recognition is typically addressed by equipping a robot with a probabilistic model that infers intent from human actions~\cite{Javdani2015,Jain2019}. Whilst the growing interest in scalable learning techniques for modelling agent intent has spurred on applications in robotics~\cite{Nicolis2018,Hu2019}, disentanglement learning remains sparse in the literature. The only known comparable work to ours is a conditional VAE that disentangled latent variables in a multi-agent driving setting~\cite{Hu2019}. Albeit similar in principle, we believe our approach is the first to infer a discrete ``intent'' variable from human behaviour by clustering action plans.
\section{Conclusions}
\label{sec:conclusions}

In this paper, we embraced an unsupervised outlook on human intention inference through a framework that disentangles and clusters latent representations of input sequences. A robotic wheelchair experiment on intention inference gleaned insights into how our proposed \discvae{} could discern primitive action plans,~\eg\ rotating in-place or reversing, without supervision. The elevated classification performance in semi-supervised learning also posits that disentanglement is a worthwhile avenue to explore in intention inference.



There are numerous promising research directions for an unsupervised means of inferring intent in human-robot interaction. The task-agnostic prior and inferred global latent variable could be exploited in long-term downstream tasks, such as user modelling, to augment the wider adoption of collaborative robotics in unconstrained environments. A truly interpretable latent structure could also prove fruitful in assistive robots that warrant explanation by visually relaying inferred intentions back to end-users~\cite{Zolotas2019}.

\bibliographystyle{IEEEtran}
\bibliography{IEEEabrv,library}

\end{document}